%% file: acl_latex.tex
\pdfoutput=1

\documentclass[11pt]{article}

\newcommand{\comment}[1]{}
\usepackage{acl}
\usepackage{svg} 
\usepackage{graphicx}
\usepackage{times}
\usepackage{latexsym}
\usepackage{amsmath}
\usepackage{tikz}
\usepackage{eso-pic}
\usepackage{booktabs}
\usepackage[T1]{fontenc}
\usepackage{multirow}
\usepackage[utf8]{inputenc}
\usepackage{amsmath, amssymb}
\usepackage{microtype}
\usepackage{xspace}

\newcommand{\ourmodel}{ReHAC\xspace}

\title{Large Language Model-based Human-Agent Collaboration \\ for Complex Task Solving}

\author{First Author \\
  Affiliation / Address line 1 \\
  Affiliation / Address line 2 \\
  Affiliation / Address line 3 \\
  \texttt{email@domain} \\\And
  Second Author \\
  Affiliation / Address line 1 \\
  Affiliation / Address line 2 \\
  Affiliation / Address line 3 \\
  \texttt{email@domain} \\}

\author{Xueyang Feng$^{1,2}$\thanks{\quad Equal Contribution. The order is determined by dice rolling.}, Zhi-Yuan Chen$^{1,2}$$^{\ast}$, Yujia Qin$^3$, Yankai Lin$^{1,2}$\thanks{\quad Corresponding Authors.}\\
\textbf{Xu Chen$^{1,2}$$^{\dagger}$, Zhiyuan Liu$^{3}$, Ji-Rong Wen$^{1,2}$}  \\
  $^1$Gaoling School of Artificial Intelligence, Renmin University of China, Beijing, China \\
  $^2$ Beijing Key Laboratory of Big Data Management and Analysis Methods, Beijing, China\\
  $^3$ Department of Computer Science and Technology, Tsinghua University, Beijing, China
  \\
    \texttt{\{xueyangfeng, zhiyuanc2001, yankailin, xu.chen\}@ruc.edu.cn} 
    }

\begin{document}
\maketitle
\begin{abstract}

In recent developments within the research community, the integration of Large Language Models (LLMs) in creating fully autonomous agents has garnered significant interest. Despite this, LLM-based agents frequently demonstrate notable shortcomings in adjusting to dynamic environments and fully grasping human needs. In this work, we introduce the problem of LLM-based human-agent collaboration for complex task-solving, exploring their synergistic potential. In addition, we propose a \underline{\textbf{Re}}inforcement Learning-based \underline{\textbf{H}}uman-\underline{\textbf{A}}gent \underline{\textbf{C}}ollaboration method, \underline{\textbf{\ourmodel}}. This approach includes a policy model designed to determine the most opportune stages for human intervention within the task-solving process. We construct a human-agent collaboration dataset to train this policy model in an offline reinforcement learning environment. Our validation tests confirm the model's effectiveness. The results demonstrate that the synergistic efforts of humans and LLM-based agents significantly improve performance in complex tasks, primarily through well-planned, limited human intervention. Datasets and code are available at: \url{https://github.com/XueyangFeng/ReHAC}.

\end{abstract}

\section{Introduction}
\label{sec:intro}
\input{sections/intro.tex}

\section{Approach}
\label{sec:approach}
\input{sections/approach.tex}

\section{Experiments}
\label{sec:experiments}
\input{sections/experiments}

\section{Discussion}
\label{sec:discussion}

\input{sections/discussion.tex}

\section{Related Work}
\label{sec:related_work}

\input{sections/related_work.tex}

\section{Conclusion}
\label{sec:conclusion}
\input{sections/conclusion.tex}

\section*{Ethical Considerations and Limitations}
The objective of this work focuses on human-agent collaboration, which requires humans to interact with LLM-based agents. We acknowledge that agents are likely to output some hallucinations and misleading information, and it is unclear how these contents impact humans. Additionally, all datasets used in this work are publicly available, and therefore, there are no data privacy concerns. All data collected will be used for research purposes only

The limitations of this paper can be summarised in three aspects:

1) The current study is confined to basic LLM-based agent architectures based on the "ReAct" and "Try Again" frameworks, while more complex architectures involving self-reflection and memory capabilities are still unexplored.  

2) Our research primarily focuses on the use of 7B and 13B scale models as policy models for task allocation. Future work will investigate the feasibility of smaller models in carrying out these tasks, aiming to maintain performance while reducing resource consumption.

3) This study is based on the assumption that human performance supersedes that of agents. However, as technology advances, agents might surpass human capabilities. Future research will thus shift towards exploring human-agent collaboration models in this new context. Emphasis will be placed on assessing how human-agent collaboration can ensure the safety of agent decisions while aligning with human preferences.

\bibliography{anthology,custom}
\bibliographystyle{acl_natbib}

\appendix

\clearpage
\section{Appendix}
\label{sec:appendix}
\input{sections/appendix}

\end{document}

%% file: sections/intro.tex
In today's increasingly complex world, humans are confronted with multifaceted tasks stemming from technical, social, and economic domains. Solving these complex tasks necessitates not only human interaction with the environment but also intricate decision-making processes. To alleviate human workload and enhance the automation of tasks in both professional and personal spheres, researchers have been actively developing advanced tools for human assistance~\citep{zawacki2019systematic, amershi2019guidelines}.
Recently, the emergence of Large Language Models (LLMs) such as LLaMA~\citep{touvron2023llama}, Gemini~\citep{team2023gemini} and GPT~\citep{brown2020language,achiam2023gpt} has marked a significant milestone. LLMs' remarkable abilities in task understanding, planning, and reasoning~\cite{zhao2023survey} have given rise to the development of LLM-based autonomous agents~\citep{wang2023survey, yao2022react, shinn2023reflexion}. These agents are designed to leverage the LLMs' capabilities to assist humans in solving complex tasks autonomously. 
The LLMs' capabilities enable them to effectively navigate and address the complexities encountered in real-world scenarios, thereby offering substantial support in human decision-making processes of task-solving.

\begin{figure}
    \centering
    \includegraphics[width=\linewidth]{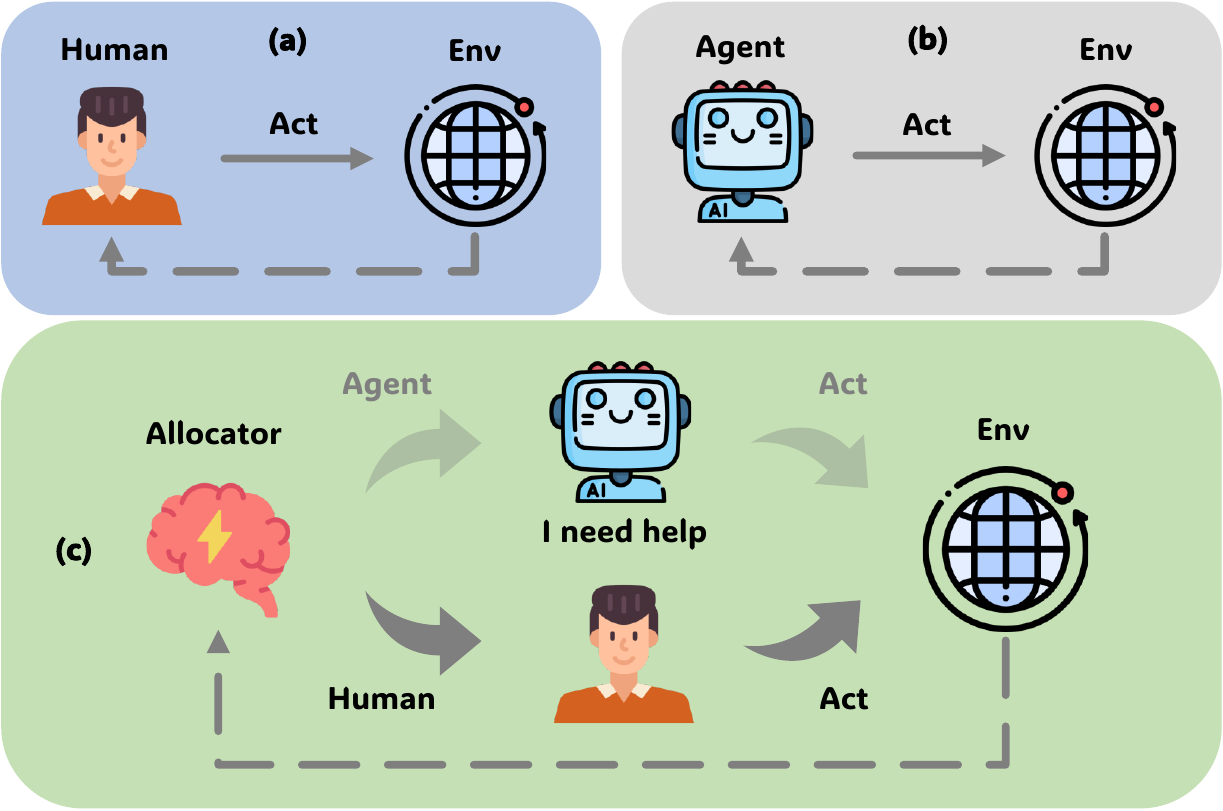}
    \caption{Different Levels of Automation. \textbf{(a) No automation:} Tasks are entirely performed by humans. \textbf{(b) Full automation:} Tasks are completely executed by agents without human intervention. \textbf{(c) Conditional automation:} Humans are required only for specific sub-tasks, without continuous monitoring.}
    \label{fig:auto}
\end{figure}

Despite the remarkable progress of LLM-based agents, there remains a notable gap in their intelligence level to handle complex and dynamic real-world tasks with human-like proficiency. This limitation poses a significant challenge to their practicality in real-world applications, especially in scenarios where high accuracy is crucial, such as the legal or financial domains. Addressing this challenge extends beyond just enhancing the agents' capabilities. Incorporating human intuition and wisdom is equally vital for the effective management of these intricate and evolving tasks, offering a complementary approach to the limitations of current agent technologies.

In this work, we introduce the problem of \textbf{LLM-based human-agent collaboration for complex task solving}, aiming to augment the capabilities of LLM-based agents by integrating human intuition and wisdom. The idea is analogous to the evolution in autonomous driving technology, which has been categorized into varying levels of autonomy, ranging from no automation, conditional automation to full automation~\cite{khan2022autonomous, sae2021automation}. Referring to this framework, we define the different levels of human-agent collaboration, as illustrated in Figure~\ref{fig:auto}.
Applying this conditional automation mode to LLM-based agents offers a practical path for their deployment in real-world scenarios, acknowledging the current limitations in their cognitive capabilities.  Instead of aiming for full automation, human-agent collaboration under the paradigm of conditional automation enables humans to intervene the complex task-solving when necessary, while agents handle most of the sub-tasks. This takes advantage of both human and machine intelligence.

While advancements in LLMs significantly enhance the capacity for mutual understanding in human-agent collaboration, several crucial challenges persist. These challenges include defining the division of labor between humans and agents, determining the granularity of tool execution, managing proactive interruption, and implementing multi-level intervention. However, our research specifically focuses on scenarios where humans directly replace agents in action.
The key challenge we aim to address in human-agent collaboration lies in determining the optimal stages for human intervention in task-solving and minimizing such intervention to enhance efficiency.
Some researchers have made preliminary attempts, by designing heuristic rules or specialized prompts to determine the stages at which agents should seek human assistance~\cite{cai2023human, wu2022ai, mehta2023improving, wang2023mint}. 
However, these rule-based or prompt-driven approaches are heavily reliant on specific application contexts and lack universality. They often demand a deep understanding of the domain and substantial experience from the designers, otherwise, suboptimal design choices can lead to reduced performance. Apart from that, a standardized formal framework and universally accepted paradigm for leveraging large language models (LLMs) in human-agent collaboration is still lacking.



To overcome the aforementioned challenges, we propose a \underline{\textbf{Re}}inforcement Learning-based \underline{\textbf{H}}uman-\underline{\textbf{A}}gent \underline{\textbf{C}}ollaboration method, \underline{\textbf{\ourmodel}}, aimed at effectively combining human intervention with the automation capabilities of LLM-based agents. Our method, leveraging reinforcement learning, trains a policy model to dynamically identify the most advantageous moments for human input during the task-solving process. \textbf{\ourmodel} is a learnable general framework that can be applied to various scenarios and does not require additional prior knowledge to design rules and prompts.
For training this policy model, we collect a dataset comprising tasks collaboratively completed by humans and LLM-based agents, utilized for the offline training of the policy model. 
We conducted extensive experiments on three multi-step reasoning datasets: HotpotQA, StrategyQA, and InterCode, using two popular LLM-based agent frameworks, ReAct and "Try-again". The experimental results indicate that with a policy model learned from limited data, \ourmodel can effectively allocate human intervention in human-agent collaboration scenarios, thereby achieving a balance between effectiveness and efficiency.

%% file: sections/approach.tex
In this section, we first formulate the problem of human-agent collaboration for complex task solving, and then introduce our proposed \ourmodel method in detail.

\subsection{Preliminary and Problem Formulation}

Complex task-solving, inherently necessitating multi-step planning and reasoning, is conventionally formalized as a multi-step decision-making problem. Historically, complex task-solving was predominantly achieved through \textbf{human-driven methods}. These methods leveraged human cognitive capabilities to determine the suitable action in each step.  Formally, considering a complex task $q$, it is traditionally solved via a sequence of actions $(a_1, a_2,  \cdots a_n)$, with each action determined by human decision-making, expressed as:
\begin{equation}
    a_t = \text{Human}(q, s_t),
\end{equation}
where $s_t = (a_1, o_1, \cdots, a_{t-1}, o_{t-1})$ denotes the history information of task state at step $t$ and $o_t$ is the observation after $a_{t-1}$ is proceeded. 

The advent of LLMs has brought a paradigm shift in this arena. Their impressive understanding and reasoning abilities have prompted research into LLM-based agents for complex task-solving, thereby enhancing the level of automation in task-solving. These \textbf{agent-driven methods} (e.g., ReAct~\citep{yao2022react}), leverage LLM-based agents to supplant human decision-making. This shift is represented as:
\begin{equation}
    a_t = \text{Agent}(q, s_t).
\end{equation}
This evolution of such AI-driven techniques provides a way to the automation of complex task-solving. 

However, limited by the current intelligence level of LLMs, full automation based on agent-driven methods is not yet feasible in practical scenarios~\citep{kiseleva2022iglu, mehta2023improving}. Inspired by autonomous driving~\citep{cui2024drive, fu2024drive, bastola2024driving}, we propose the problem of \textbf{LLM-based human-agent collaboration for complex task solving} and explore the dynamics and efficacy of the \textbf{human-agent collaborative methods} for complex task solving. We first explore a specific form of  human-agent collaboration: humans intervene in the complex task-solving process when necessary. Formally, we need to determine whether a human or an agent makes decisions based on the actions' complexity and contextual changes, i.e., 
\begin{equation}
    a_t = \text{Human}(q, s_t) \quad\text{or} \quad\text{Agent}(q, s_t),
\end{equation}

It is generally perceived that direct human intervention in decision-making, particularly in real-world scenarios, incurs higher costs and diminishes the system's automation level~\citep{cai2023human, wang2023mint}. On the other hand, human intervention plays an important role in enhancing task performance and flexibility. Therefore, the objective of human-agent collaboration is to enhance the effectiveness of complex task-solving with minimal reliance on human decision-making.
One key challenge is to \textbf{determine the stages in the task-solving process where human intervention is most beneficial and effective, aligning with the goal of minimizing human involvement while maximizing task performance}.

\subsection{\ourmodel}

In this work, we propose a Reinforcement learning-based Human-Agent Collaboration method, \ourmodel. It formulates the human-agent collaboration problem as a Markov Decision Process (MDP) framework, represented by the tuple $(S, \mathcal{A}, P, R, \gamma)$, where $S$ is the set of states, $\mathcal{A}$ is the set of actions, $P: S\times \mathcal{A}\times S$ is the state transition probabilities, $R$ serves as the reward function, and $\gamma$ the discount factor. 

For each action $a_t\in \mathcal{A}$, we define it as a tuple $(a^{collab}_t, a^{task}_t)$, where $a^{collab}_t$ indicates the subtask is allocated to an agent or a human, and $a^{task}_t$ is the task action determined by agent or human:
\begin{equation}
    a^{collab}_t \sim \pi^{collab}_{\theta_1}(a^{collab}_t|s_t) \nonumber
\end{equation}
\begin{equation}
    a^{task}_t \sim  
    \begin{cases} 
        & \pi^{task}_{\theta_2}(a^{task}_t|s_t), \quad \text{if}\ \  a^{collab}_t = 0;  \\
        &\pi^{task}_{\text{Human}}(a^{task}_t|s_t), \quad \text{otherwise},
    \end{cases}
\end{equation}
where $\pi^{collab}_{\theta_1}$ is the collaboration policy model, $\pi^{task}_{\theta_2}$ is the agent-based task policy model, and $\pi^{task}_{\text{Human}}$ is the human task policy. 

To balance the maximization of task performance and the cost of human intervention, we define the reward function as:
\begin{equation}
\label{equ:reward_compute}
R(s, a) = T(s, a) - \lambda C(s, a),
\end{equation}
where $T(s, a)$ is the measure of expected task rewards received after taking action $a$ in state $s$, $C(s, a)$ is the number of human interventions in the trajectory after taking action $a$, $\lambda$ is a hyper-parameter that serves as a penalty coefficient of the number of human interventions. We utilize Monte-Carlo estimation to compute this reward function.

\paragraph{Optimization:} Following the REINFORCE algorithm \citep{williams1992simple}, we optimize the expected reward:
\begin{equation}
    \mathcal{J}(\pi_{\theta}) = \mathbb{E}_{\pi_{\theta}}[R(s, a)],
\end{equation}
which aims to find an optimal policy $\pi_\theta$ that ensures the maximization of task rewards while minimizing the human intervention costs, and $\theta = [\theta_1, \theta_2]$. We utilize the advantage function to enhance the stability of optimization and important sampling for offline learning:
\begin{gather}
    A(s, a) = R(s, a) - \frac{1}{|\mathcal{A}|} \sum_{a' \in \mathcal{A}} {R(s, a')}
    \nonumber\\
    \nabla_{\theta} \mathcal{J}(\pi_\theta) = \sum_{s} \sum_{a} w(s, a) \nabla_{\theta} \log \pi_{\theta}(a | s) A(s, a), 
    \nonumber\\
    w(h, a) = \text{Clip}\left(\frac{\pi_\theta(s, a)}{\pi_{\text{beh}}(s, a)}\right),
\end{gather}
where $A(s, a)$ is the advantage function, the clip function limits the importance sampling term to the interval \({1-\epsilon}\) to \({1+\epsilon}\), and the behavior policy $\pi_{\text{beh}}$ represents the policy under of the offline training. Moreover, we have incorporated an entropy regularization term. This term encourages the policy to explore a variety of actions, thereby preventing the policy from becoming too deterministic and overfitting to the training data. Finally, the gradient of objective function is as follows:
\begin{equation}
    \nabla_{\theta} \tilde{\mathcal{J}}(\pi_\theta) = \nabla_{\theta} \mathcal{J}(\pi_\theta) + \alpha \nabla_{\theta} H(\pi_\theta(\cdot | s)).
\end{equation}

%% file: sections/experiments.tex
\begin{figure*}[!ht]
    \centering
    \includegraphics[width=0.8\linewidth]{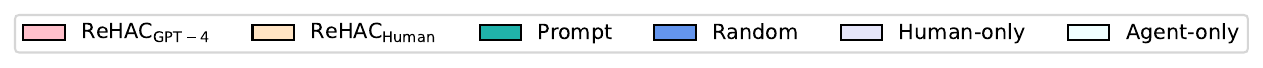}

    \begin{minipage}{0.49\textwidth}
        \centering
        \includegraphics[width=\textwidth]{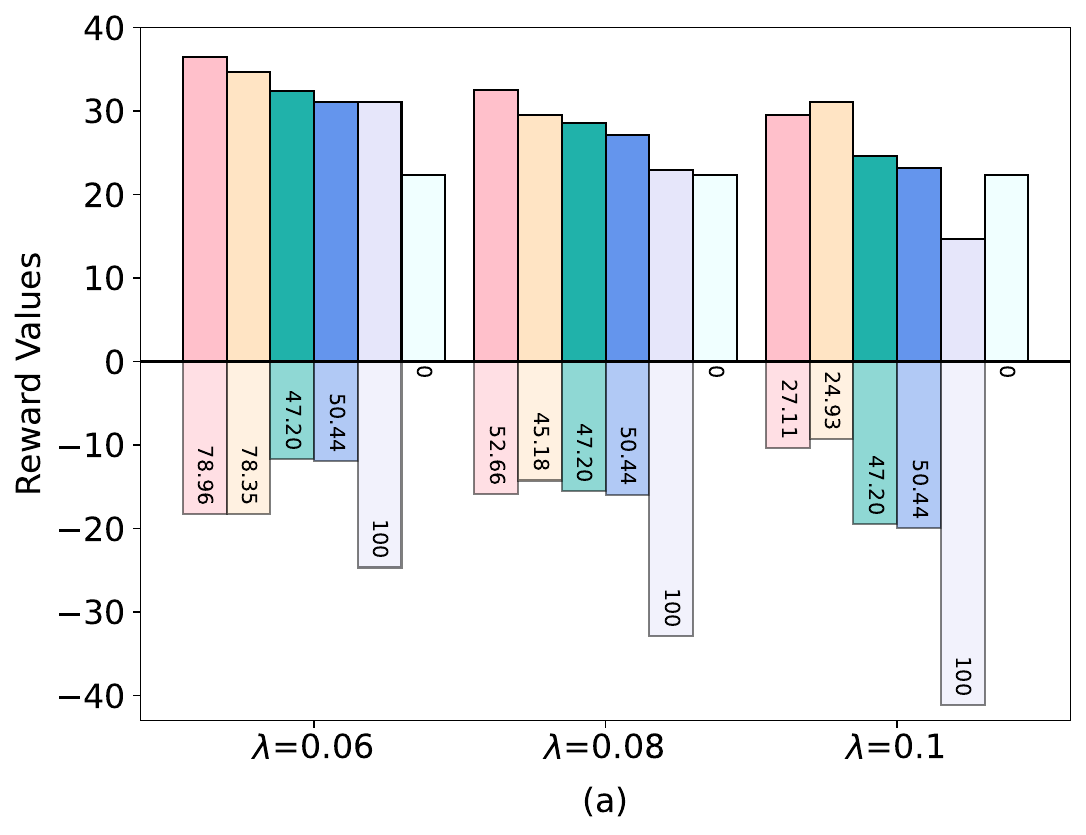} 
    \end{minipage}\hfill
    \begin{minipage}{0.49\textwidth}
        \centering
        \includegraphics[width=\textwidth]{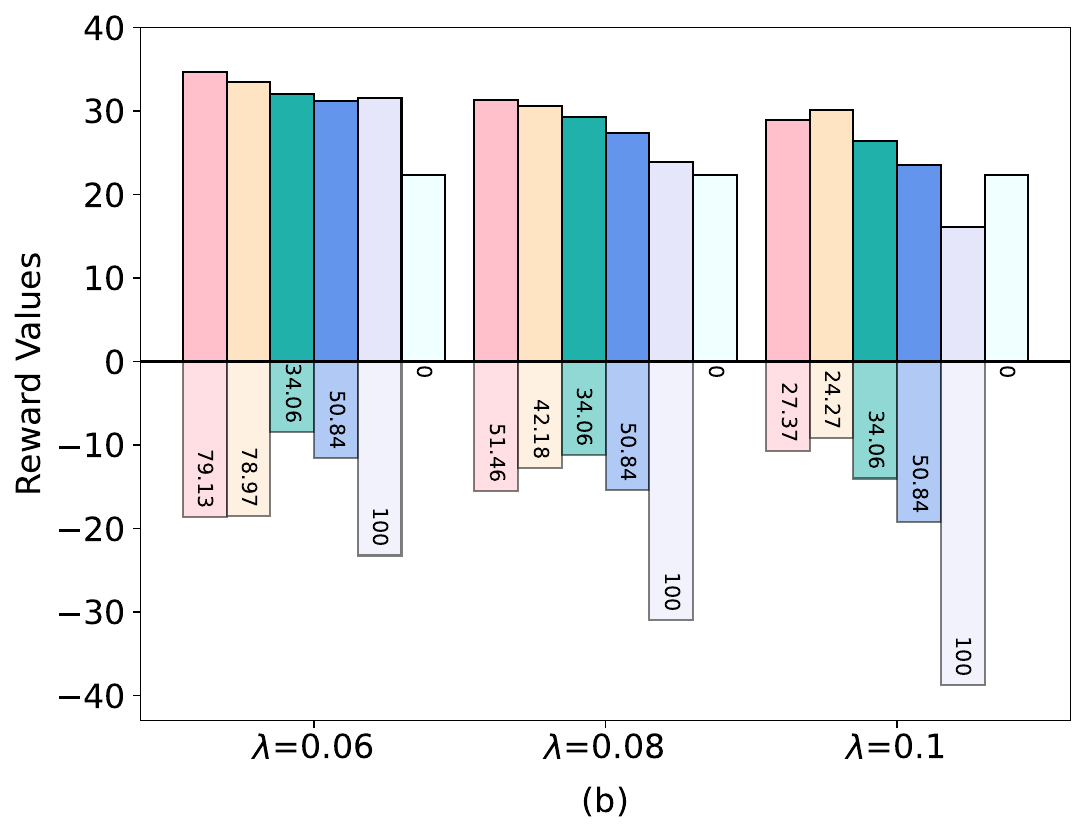} 
    \end{minipage}
    \caption{(a) Human-agent collaboration evaluation. (b) GPT-4-agent collaboration evaluation.
    The bars above the 0-axis represent the reward $R$, the bars below the 0-axis represent the human intervention cost $\lambda C$, and the entire columns, composed of the bars above and below the 0-axis, represent the task reward $T$. Numbers within the bars means the human intervention rate (\%). $\text{\ourmodel}_{\text{GPT-4}}$ and $\text{\ourmodel}_{\text{Human}}$ represent the policy model trained on GPT-4-agent and human-agent collaboration datasets, respectively.
    \ourmodel outperforms other baselines in human-agent collaboration scenarios.}
    \label{fig:overall-result}
\end{figure*}

\subsection{Experimental Setup}

\input{sections/experimental_setups}

\subsection{Overall Results}

In this section, we verify the effectiveness of our proposed \ourmodel method for human-agent collaboration on the HotpotQA dataset. 

\paragraph{Human-Agent Experiments}~Figure~\ref{fig:overall-result}(a) shows the evaluation results of human-agent collaboration on the HotpotQA dataset. 
From the figure, we can observe that all human-agent collaboration methods outperform Human-only and Agent-only methods. This underscores the importance of collaborating human and agent in complex task-solving for getting higher reward.
In addition, $\text{\ourmodel}_{\text{Human}}$ achieves the best performance compared with prompt-based and random-based method in achieving higher rewards.
Specifically, when $\lambda=0.06$, \ourmodel achieves a higher reward with approximately $30\%$ more human interventions compared with the prompt-based baseline; when $\lambda=0.1$, it also achieves a reward improvement with about $20\%$ less human interventions.
This indicates that our \ourmodel method can dynamically introduce human intervention in real human-agent collaboration scenarios, thereby achieving a balance between effectiveness and efficiency.

Focusing on $\text{\ourmodel}_{\text{Human}}$, we observe that as $\lambda$ increases, the human intervention rate\footnote{The formula for calculating the human intervention rate is in Appendix \ref{human_intervnetion_rate_formula}.} (HIR) of $\text{\ourmodel}_{\text{Human}}$ gradually decreases.
This trend suggests that a higher human penalty coefficient elevates our policy model's ``threshold'' for assigning actions to humans. Simultaneously, the decrease of the HIR correspondingly results in a deterioration of human-agent interaction performance.

\paragraph{Human Simulation}~Due to the high cost of hiring annotators to label real human-agent collaboration data, it is costly for us to collect human-agent collaboration data on more datasets and, as a result, validate the efficacy of our method in broader scenarios.
We instead use GPT-4 (gpt-4-0613) to build a simulation environment and make it collaborate with agents to solve tasks. 
This setup enables us to collect more ``human-agent'' collaboration data at a reasonable cost.

To verify the feasibility of using GPT-4 to simulate humans to collect ``human-agent'' collaboration data, we learn \ourmodel on the HotpotQA GPT-4-agent collaboration data, named as $\text{\ourmodel}_{\text{GPT-4}}$ and test its performance in the real human-agent collaboration environment.
From Figure~\ref{fig:overall-result}(a), we can see that $\text{\ourmodel}_{\text{GPT-4}}$ exhibits better performance compared to $\text{\ourmodel}_{\text{Human}}$ in human-agent collaboration when $\lambda=0.06\text{ and }0.08$. We suppose that this is possibly attributed to individual differences among humans, leading to a distribution variance in the human-agent collaboration data, while GPT-4-agent collaboration data exhibits higher consistency and lower variance. This makes $\text{\ourmodel}_{\text{GPT-4}}$ learn the collaboration signal more easily, and thus is more stable and performs better.

To further reduce costs and observe the reward variation of \ourmodel during the training process, we use GPT-4 to simulate humans in the evaluation phase.
Figure~\ref{fig:overall-result}(b) shows the evaluation results when using GPT-4 to simulate humans for collaboration. Comparing the results in Figure~\ref{fig:overall-result}(a) and (b), we notice that the relative performance of various methods is generally consistent in both human-agent collaboration and GPT-4-agent collaboration.
For example, the rewards $R$ of \ourmodel consistently surpass those of the Prompt method, and both \ourmodel and the Prompt method outperform the Random method.
This demonstrates the viability of using GPT-4 to simulate humans for evaluation.

Considering feasibility and cost-effectiveness, we will continue to use GPT-4 as a substitute for human participants in all subsequent extension experiments.

\begin{figure*}[!t]
    \centering
    \includegraphics[width=0.7\linewidth]{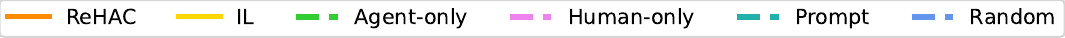}

    \begin{minipage}{0.325\linewidth}
        \centering
        \includegraphics[width=1\linewidth]{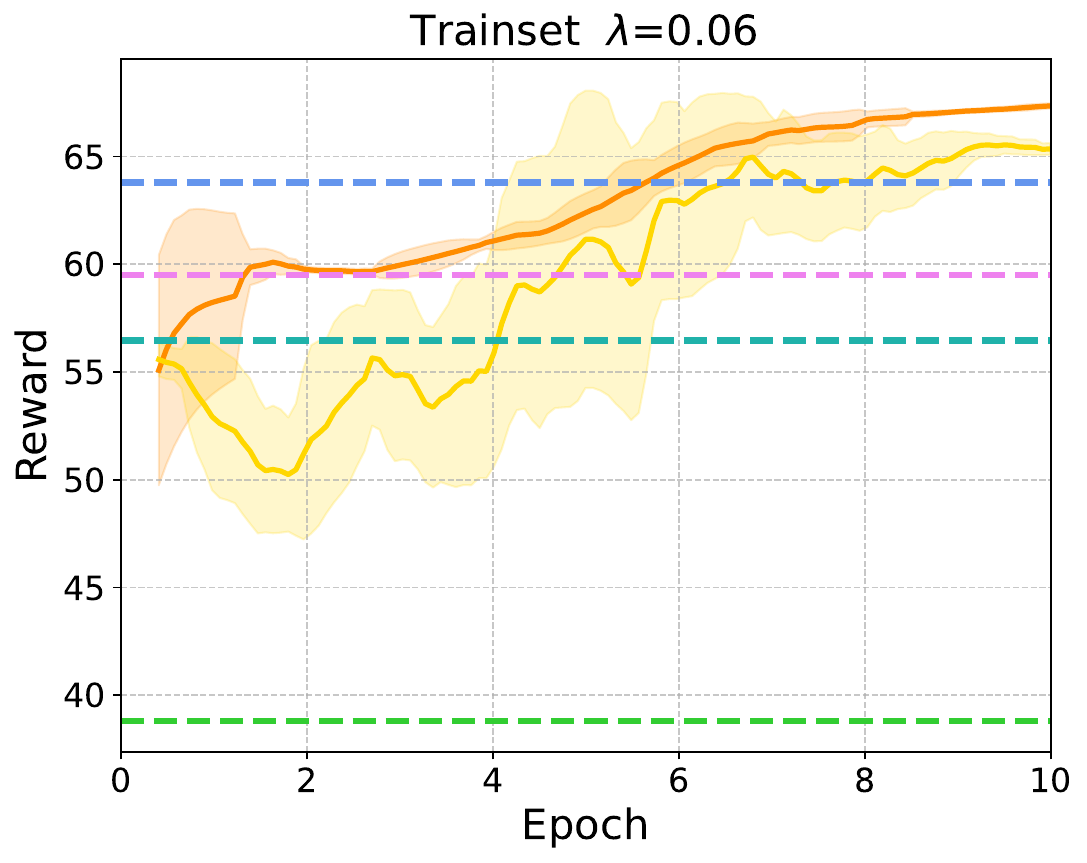}
    \end{minipage}
    \begin{minipage}{0.325\linewidth}
        \centering
        \includegraphics[width=1\linewidth]{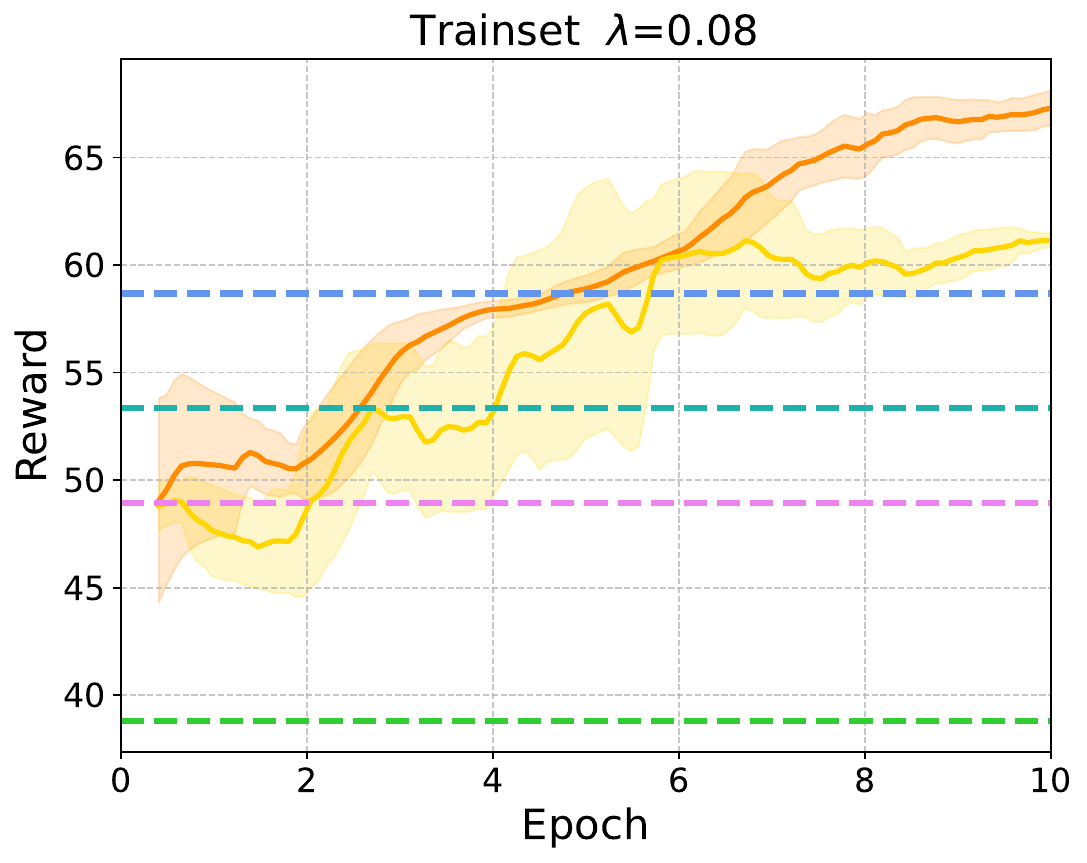}
    \end{minipage}
    \begin{minipage}{0.325\linewidth}
        \centering
        \includegraphics[width=1\linewidth]{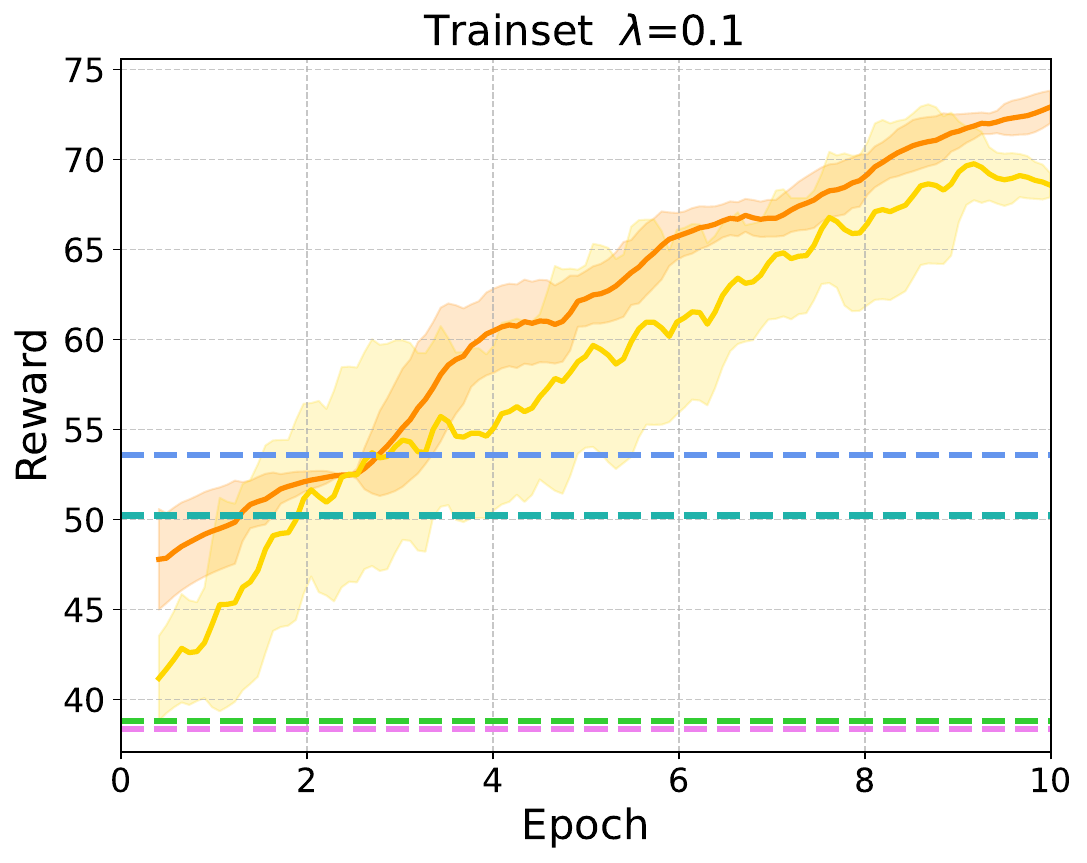}
    \end{minipage}

    \begin{minipage}{0.325\linewidth}
        \centering
        \includegraphics[width=1\linewidth]{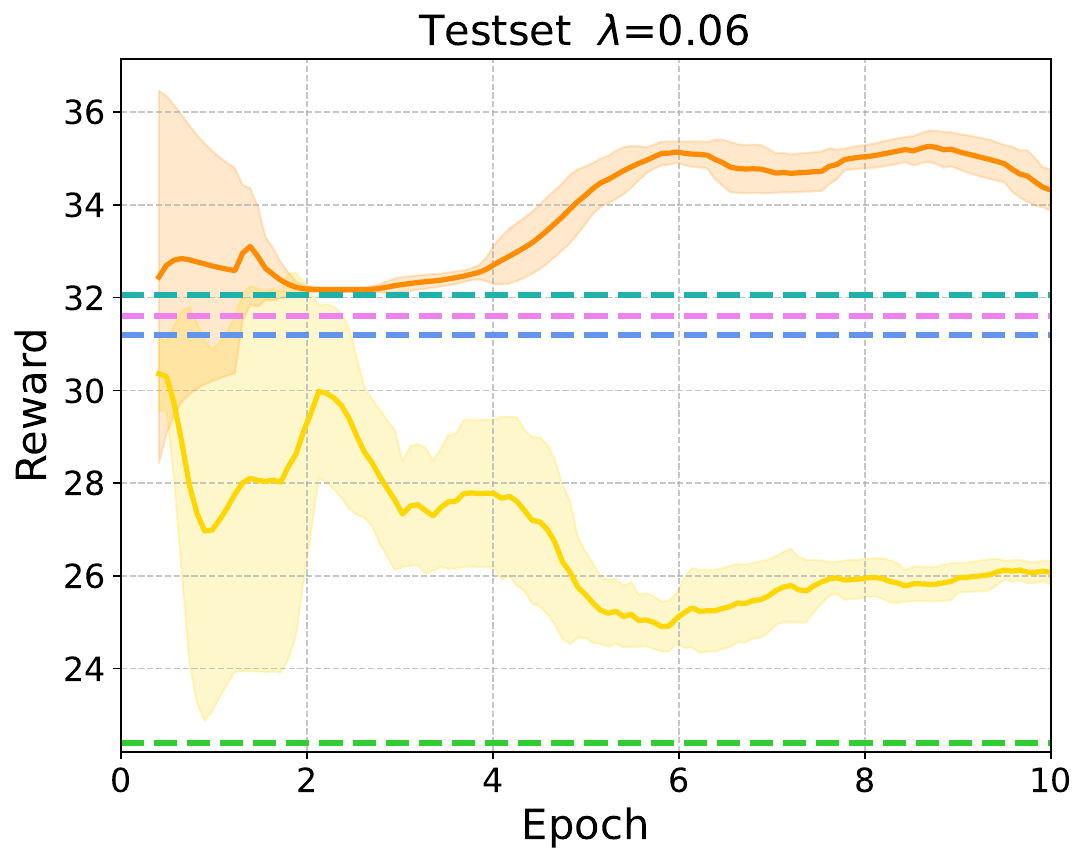}
    \end{minipage}
    \begin{minipage}{0.325\linewidth}
        \centering
        \includegraphics[width=1\linewidth]{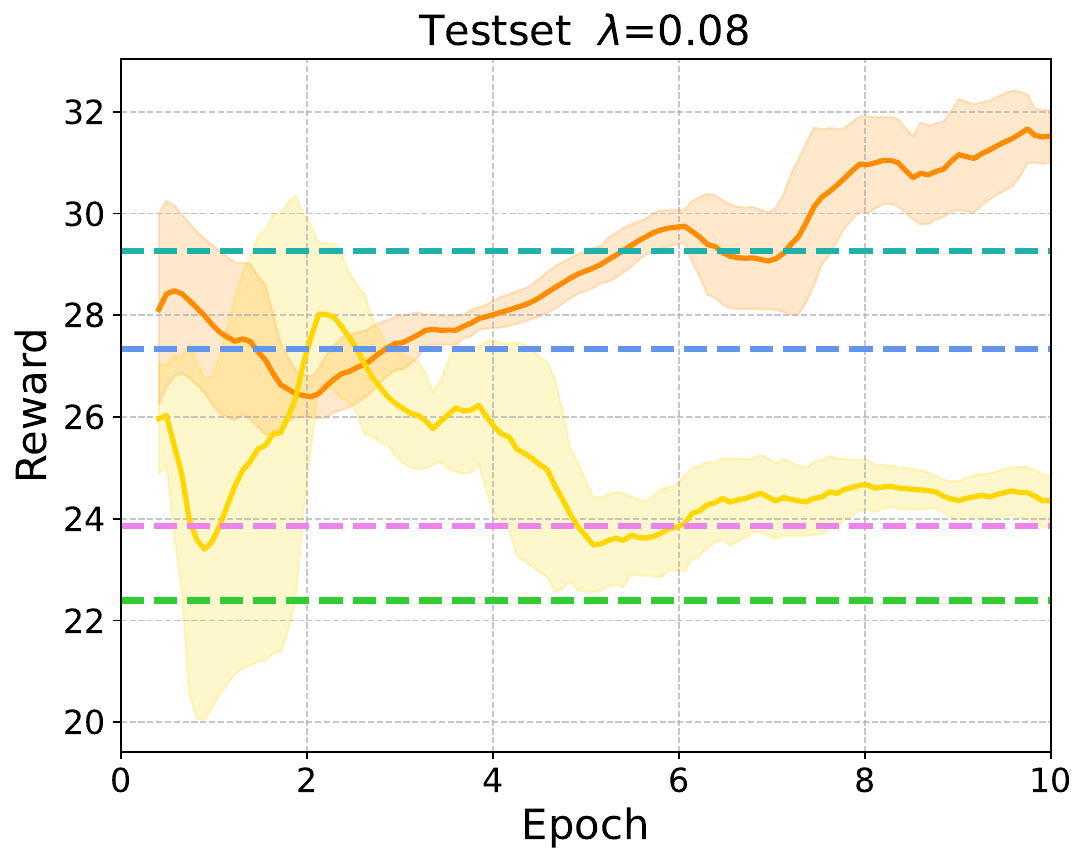}
    \end{minipage}
    \begin{minipage}{0.325\linewidth}
        \centering
        \includegraphics[width=1\linewidth]{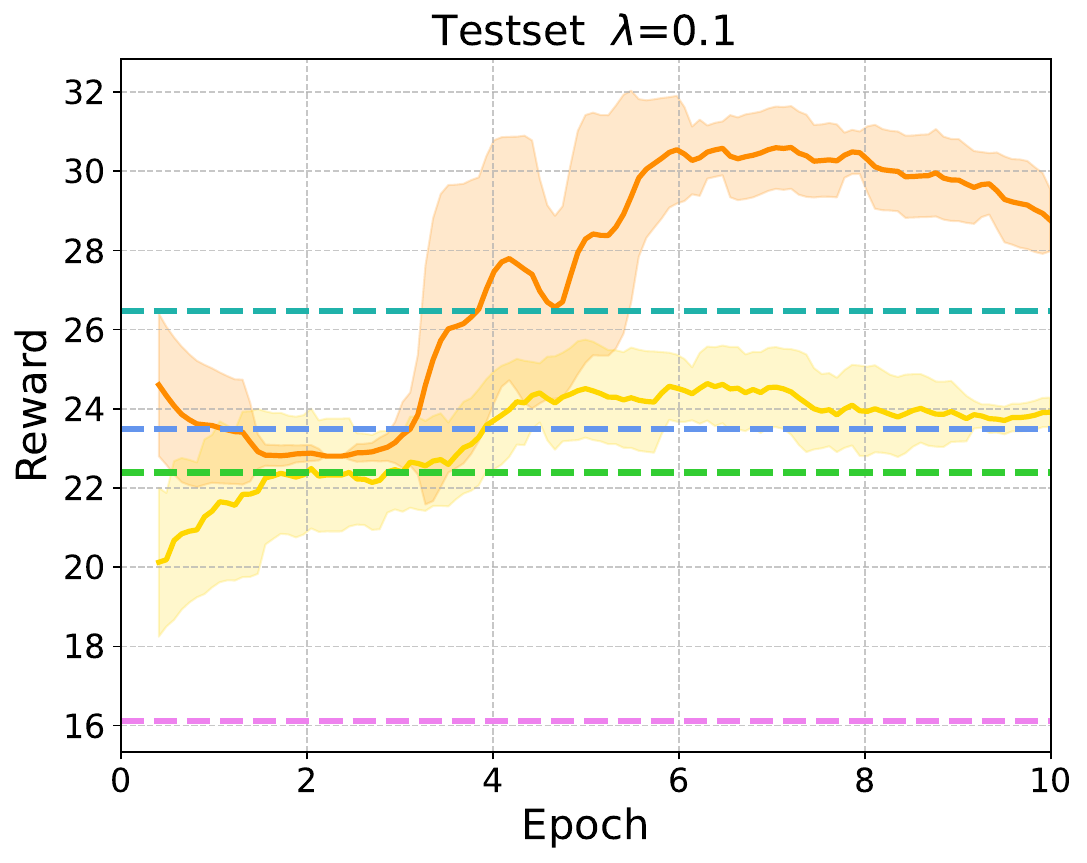}
    \end{minipage}
    
    \caption{Reward $R$ variations of different methods during the training process on HotpotQA dataset. Here we set the human intervention penalty coefficient $\lambda$ to 0.06, 0.08, and 0.1. Curves of \ourmodel and IL are averaged over 15 points, with shadows indicating the variance.}
    \label{fig:para_sensitivity}
\end{figure*}

\begin{figure*}[!t]
    \centering
    \includegraphics[width=0.7\linewidth]{pic/legend.pdf}

    \begin{minipage}{0.24\linewidth}
        \centering
        \includegraphics[width=1\linewidth]{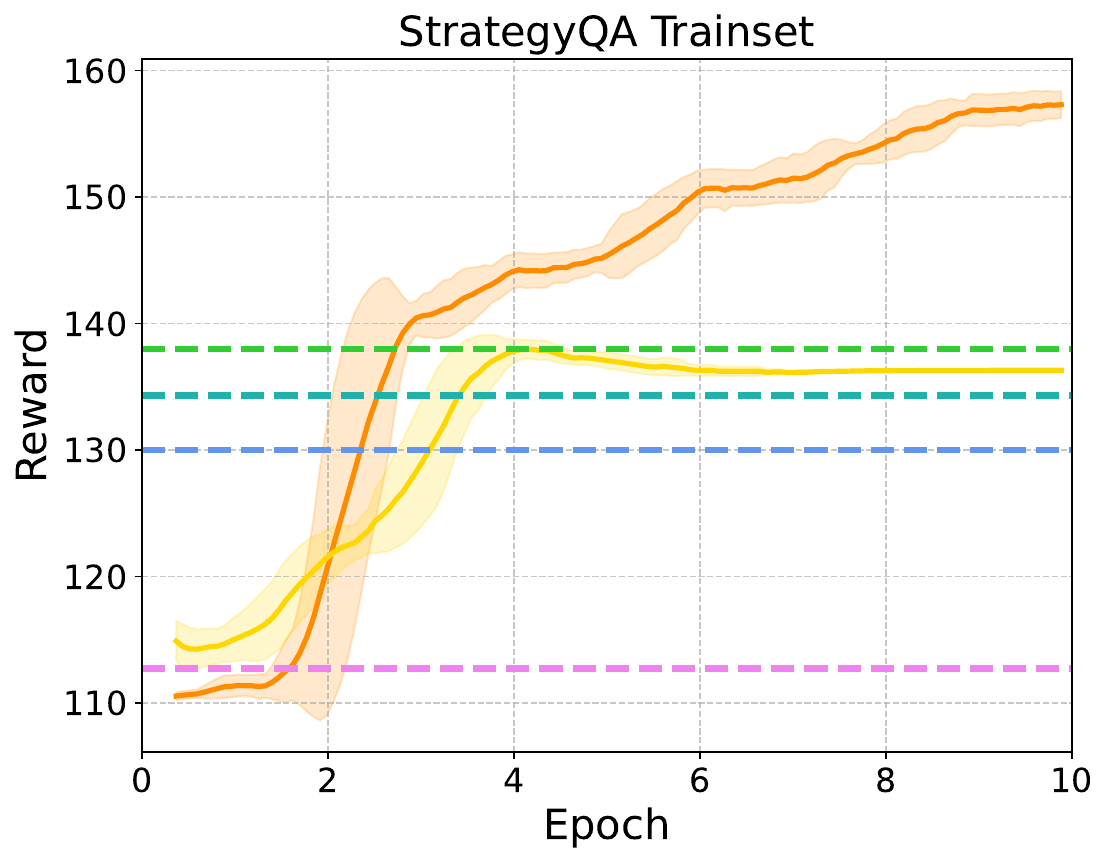}
    \end{minipage}
    \begin{minipage}{0.24\linewidth}
        \centering
        \includegraphics[width=1\linewidth]{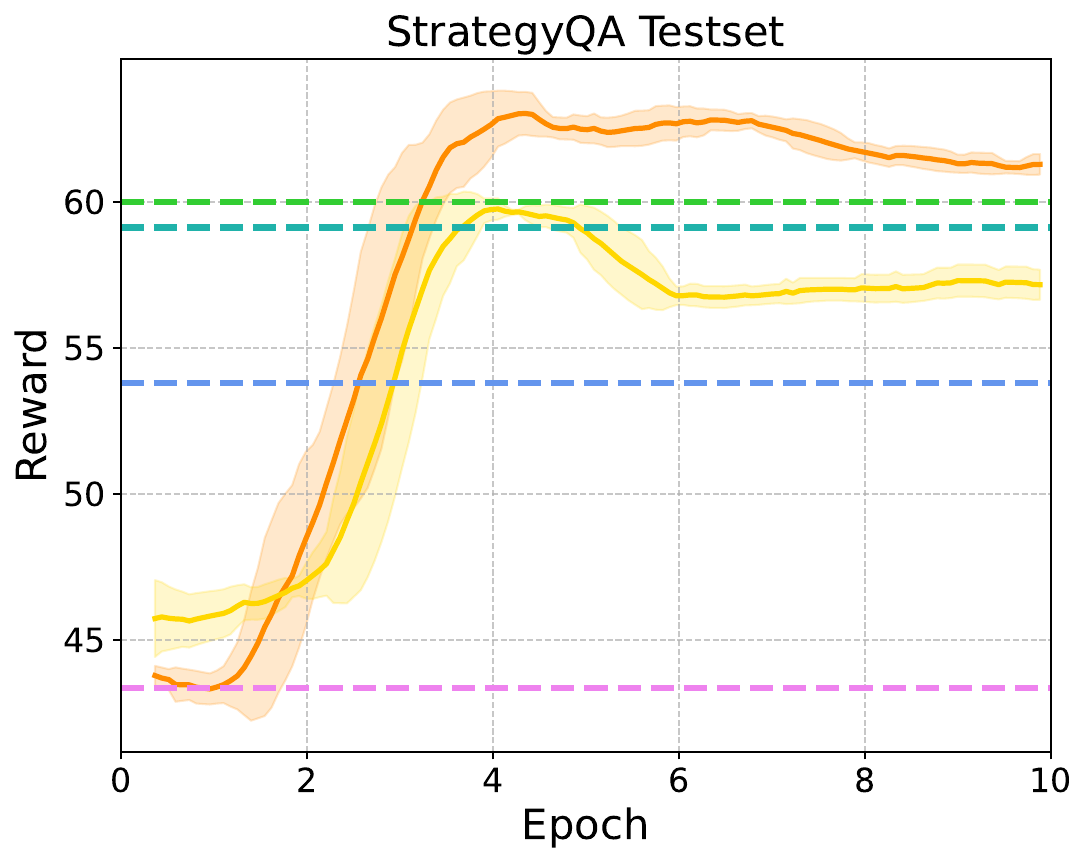}
    \end{minipage}
    \begin{minipage}{0.24\linewidth}
        \centering
        \includegraphics[width=1\linewidth]{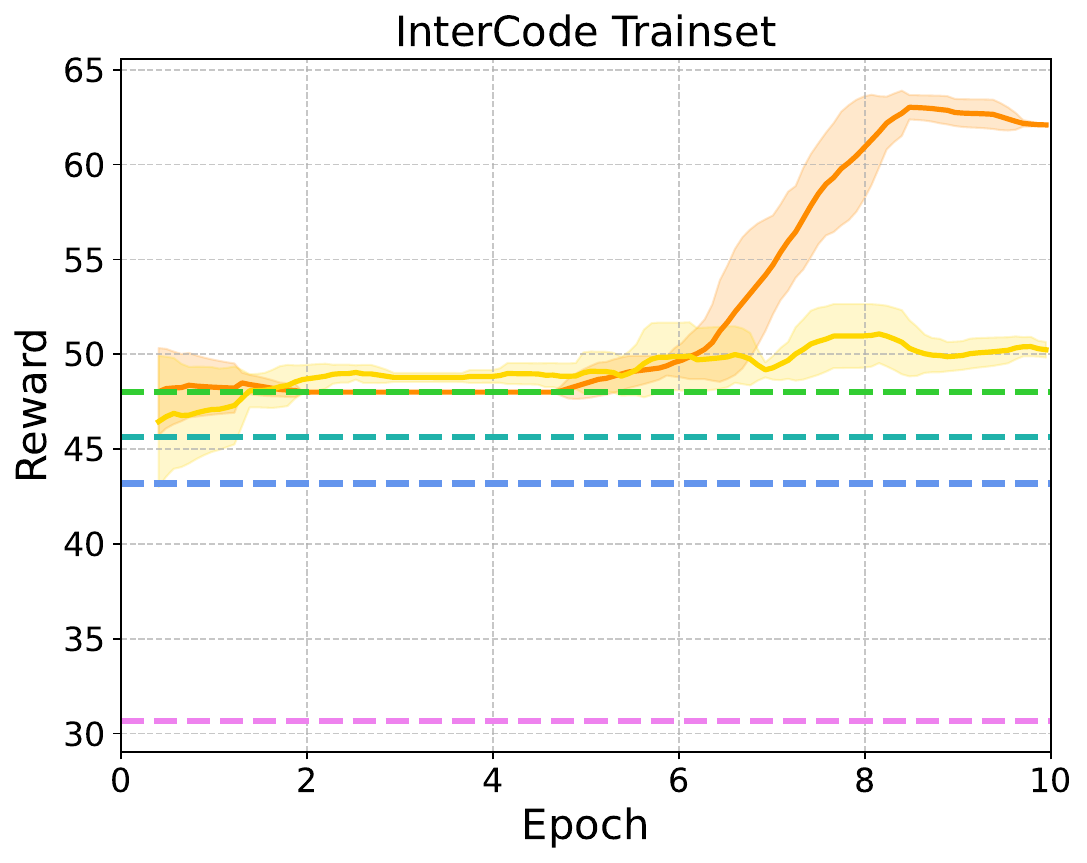}
    \end{minipage}
    \begin{minipage}{0.24\linewidth}
        \centering
        \includegraphics[width=1\linewidth]{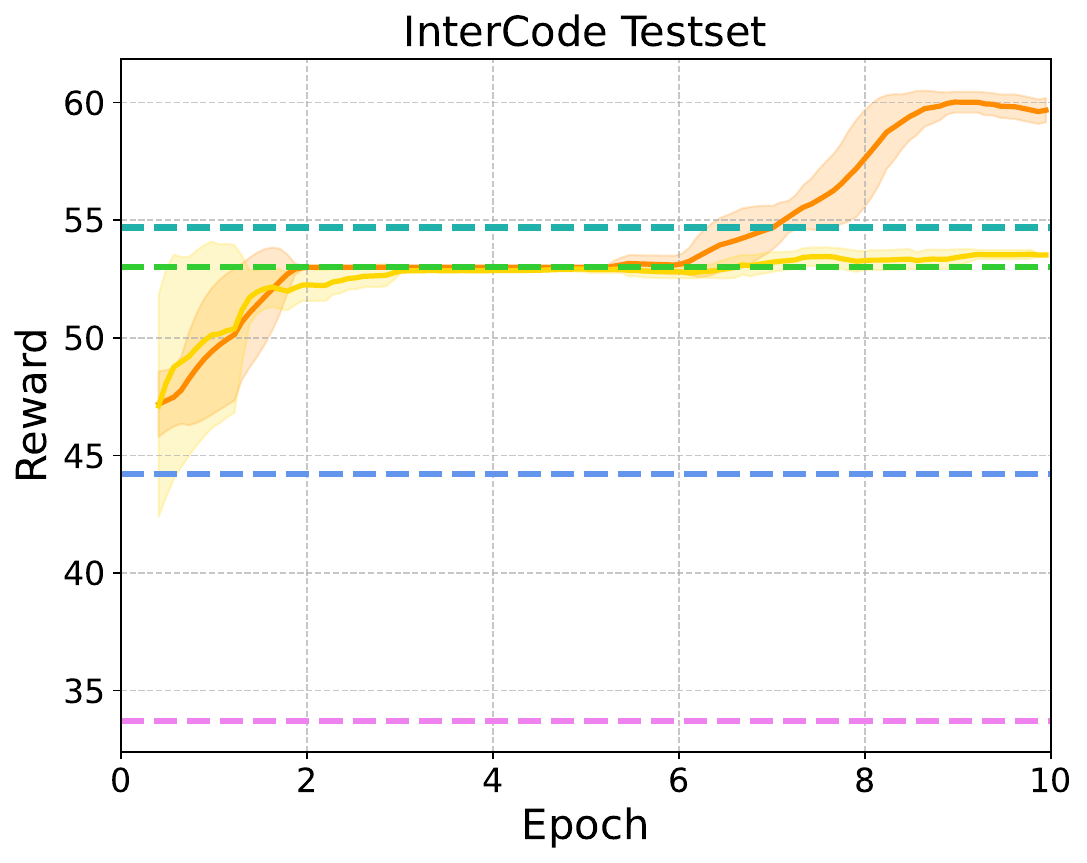}
    \end{minipage}
    
    \caption{Reward $R$ variations during the training process on three datasets. Curves of \ourmodel and IL are averaged over 15 points, with shadows indicating the variance.}
    \label{fig:gpt4_main}
\end{figure*}

\paragraph{Learning Curves} Figure~\ref{fig:para_sensitivity} shows the learning curves during the training process. 
The curves are obtained by assessing the policy model's rewards on the trainset and testset every 5 steps.
From the figure, we can observe that (1) the rewards of \ourmodel gradually increase during the training process, indicating that \ourmodel can progressively identify suitable points to introduce human interventions. (2) 
While the IL method achieves high rewards on the trainset, it performs poorly on the testset. This suggests our RL-based learning method learns a more generalized human-agent collaboration strategy compared to directly learning the optimal strategy with the imitation learning method.

\subsection{Performance on Different Dataset}
In this part, we train and test \ourmodel method on StrategyQA, and InterCode datasets in the GPT-4 simulation environment.
For all experiments, we fix the parameter $\lambda=0.08$. 
Throughout the training phase, we evaluate the policy model's rewards on the trainset and testset every 5 steps. 
Experimental results are shown in Figure~\ref{fig:gpt4_main}. From the figure, we observe that: (1) Our proposed \ourmodel method achieves higher reward scores compared to other baselines on all datasets. This validates the effectiveness of our approach across a broader range of datasets.
(2) Both \ourmodel and IL exhibit low variance and stability during the training process. 
Although our method and the IL method show a continuous reward increase during the training process, \ourmodel can ultimately achieve higher rewards compared to the IL method.
This indicates that our reinforcement learning-based method can provide more valuable guidance to the policy model $\pi_{\theta_1}^{collab}$, enabling it to determine when to introduce human interventions and consequently achieving higher rewards.

In summary, our method demonstrates superior performance across all datasets, affirming its ability to achieve an optimal balance between efficiency and effectiveness.

\label{sec:model_scale}
\begin{table}[t]
    \centering
    \small
    \setlength\tabcolsep{1pt}
    \begin{tabular}{llrcc}
    \toprule
         Dataset & Model & HIR (\%) & Task Reward $T$ & Reward $R$\\
         \midrule
         \multirow{2}{*}{HotpotQA} & LLaMA-7B & 51.46 & 46.90 & 31.38 \\
         & LLaMA-13B & 47.64 & 46.78 & 32.22 \\
         \midrule
         \multirow{2}{*}{InterCode} & LLaMA-7B & 4.15 & 62.00 & 60.08 \\
         & LLaMA-13B & 3.10 & 60.00 & 58.56 \\
    \bottomrule
    \end{tabular}
    \caption{Experimental results regarding different model scales. HIR represents the human intervention rate.}
    \label{tab:model_size}
\end{table}

\subsection{Scaling Analysis of Policy Model}
In this section, we analyze the impact of the model scale on the performance of the policy model.
Here, we set $\lambda=0.08$ and conduct experiments on HotpotQA and InterCode datasets.
As shown in Table \ref{tab:model_size}, the LLaMA-7B model performs competitively with the LLaMA-13B model.
This suggests that the Llama2-7B model is already proficient in handling the human-agent collaboration task, and the benefit of increasing the size of the model is slight. We will explore smaller policy model size in the future.

\begin{figure}[t]
    \centering
    \includegraphics[width=\linewidth]{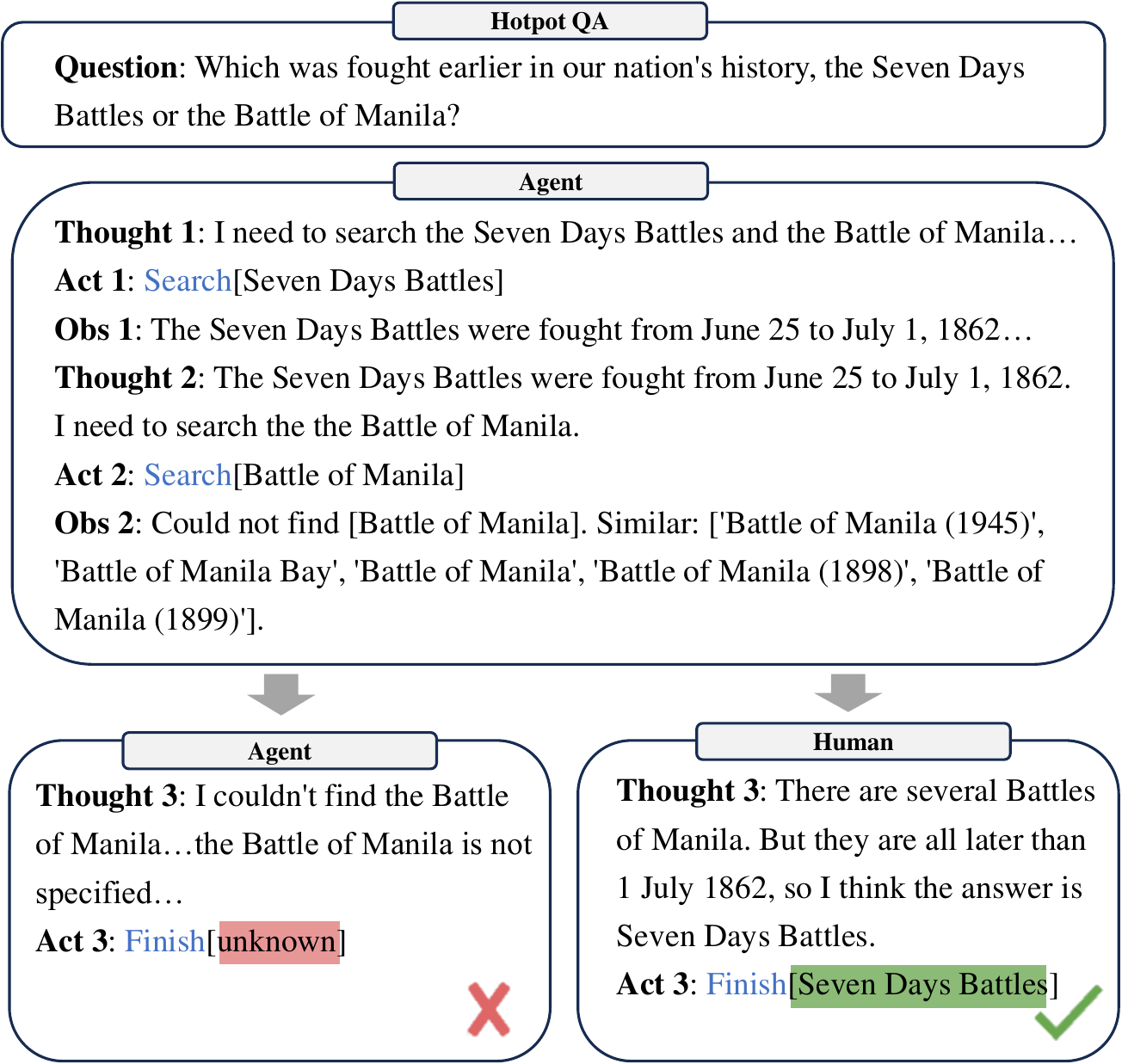}
    \caption{Case Study. When the agent completes the task, the third step cannot be answered due to the ambiguity of the problem identified; using our method, the first two simple retrieval tasks are assigned to the agent to complete, while the third step is assigned to humans. Humans can complete the correct answer through bold speculation}
    \label{fig:case_study}
\end{figure}

\subsection{Case Study}
In this part, we give a specific case on the HotpotQA dataset, as illustrated in Figure~\ref{fig:case_study}, to show how human-agent collaboration helps the complex task-solving. The task is to determine which historical event, the Seven Days Battles or the Battle of Manila, occurred first. When given the entire problem, the agent accurately determines the date of the Seven Days Battles but encounters multiple entries for the Battle of Manila, resulting in ambiguity. Consequently, the agent deems the query ambiguous and opts to respond with ``unknown''. On the contrary, our \ourmodel method requires the human intervention in this situation.
Upon examining the related entries, the human observes that all mentioned dates for the Battle of Manila occurs after to July 1, 1862. Based on this insight, he conjectures that the Seven Days Battles occurred first. Although this conjecture is not absolutely certain, it represents the most likely decision based on the available information. Thus, our \ourmodel method returns a correct response ``Seven Days Battles''. This case also highlights an insightful aspect of our research into LLM-based agents: 
Researchers are committed to eliminating hallucinations in large language models (LLMs) to create rigorous and accurate intelligent agents. However, many tasks require imagination and intuition, making it crucial to integrate human creative thinking through human-agent collaboration at this juncture.

%% file: sections/experimental_setups.tex
\paragraph{Datasets}
Following \citet{yao2022react, shinn2023reflexion, liu2023agentbench, xu2023lemur}, we evaluate the efficacy of our method on question answering and coding datasets: (1) HotpotQA \citep{yang2018hotpotqa} is a Wikipedia-based question answering benchmark which needs model to perform multi-hop reasoning over complex questions. 
(2) StrategyQA \citep{geva2021did} is a question answering benchmark with questions that need implicit reasoning. 
(3) InterCode \citep{yang2023intercode} is an interactive coding dataset that enables agents to receive feedback from the code interpreter. In this work, we use InterCode-SQL part, which requires models to write SQL statements to fulfil the query.

\paragraph{Implementation details}~We use LLaMA-2 \citep{touvron2023llama} as the collaboration policy model $\pi_{\theta_1}^{collab}$ and use Low-Rank Adaptation (LoRA, \citet{hu2021lora}) methods to train the policy model. In all experiments, we utilized ChatGPT (gpt-3.5-turbo-0613) to simulate the agent policy $\pi_{\theta_2}^{task}$. More model implementation and data collection details can be found in Appendix~\ref{implementation deltails appendix}.

In this study, we set humans and agents to solve tasks under the ReAct framework \citep{yao2022react} for question-answering datasets.
The action space of $a^{task}$ is \{{Search}[entity], {Lookup}[keyword], and {Finish}[answer]\}. All actions are supported by a Wikipedia web API, following the original ReAct implementation.
For the InterCode dataset, we solve tasks under the ``Try Again'' framework \citep{yang2023intercode}. Here, agents and humans interact with the code interpreter through the action $a_t$ and receive execution outputs from the code interpreter as observations $o_t$. The task-solving process ends if any one of the following conditions is satisfied: 1) the Finish[answer] action is executed actively by $\pi^{task}_{\theta_2}$ for the question answering dataset. 2) the task reward $T(s, a)=1$ for InterCode dataset. 3) the number of actions $t$ exceeds a pre-defined step threshold.

\paragraph{Reward Calculation}~For all datasets, the final reward is computed as equation (\ref{equ:reward_compute}).
For question answering datasets, we choose the F1 score
as the task reward \(T(s, a)\). For the InterCode dataset, following \citet{yang2023intercode}, we use Intersection over Union as the task reward $T(s, a)$.

\paragraph{Baselines}~We compare our method \ourmodel with the following baselines: 1) Agent-only which carries out all actions by agents. 2) Human-only, which conducts all actions by humans. 3) Random, which selects an agent or human randomly at a probability of 50\% to perform each action. 4) Prompt, which prompts the agent to actively decide whether the action is executed by itself or a human. 5) Imitation Learning (IL), which trains the policy model to decide whether the action should be finished by an agent or human by the IL method. More details about baselines can be found in the Appendix~\ref{baselines deltails appendix}.

%% file: sections/discussion.tex
In this paper, we conduct a preliminary exploration of key aspects of human-agent collaboration, aiming to lay the groundwork for further research in this field. Despite progress, unresolved problems and potential challenges persist. We propose three extended research directions to enhance the effectiveness, safety, and intelligence of human-agent collaboration:

\paragraph{Multi-level Human-Agent Collaboration}~Our focus is on modes where humans directly replace agents in action. However, given the distinct advantages of both humans and agents, we see a need to explore more complex collaboration levels. This includes human involvement in feedback, decision modification, and planning. 

\paragraph{Development Stages of LLM-based Agents}~Inspired by the L1 to L5 grading model in autonomous driving, we suggest adapting this framework for LLM-based human-agent collaboration. It offers a clear structure to assess the current development stage of human-agent technologies and guide future research. While LLM agents have not reached high or full automation, this framework is crucial for identifying key technologies and challenges. However, our research indicates a significant gap before LLM agents achieve full automation (L5). Effective human-agent collaboration could be a bridge towards this goal.

\paragraph{Safety and Super Alignment}~Safety is paramount in human-agent collaboration, particularly in high-risk scenarios. It's vital to explore methods to secure the collaboration process and mitigate risks. Moreover, with the potential of LLM-based agents evolving into superintelligence, effective collaboration becomes increasingly crucial. This collaboration is key, as it not only allows humans to guide ethical and safety decisions but also ensures the alignment of LLM-based agents' objectives with human interests.

%% file: sections/related_work.tex
\paragraph{LLM-based Agent}~Recent advancements in LLMs have demonstrated their capabilities in reasoning~\citep{jason2022chain, kojima2022large, hao2023reasoning, luong2024reft, yue2023mammoth} and task planning~\citep{yao2023tree, kong2023tptu2, shen2023hugginggpt, yao2023retroformer, deng2023plug}. These capabilities lay the foundation for the development of LLM-based agents~\citep{Mohit2021ALFWorld, yang2023intercode, liu2023agentbench, song2023llmplanner, wang2023survey}. LLM-based agents, which can interact with the environment and select subsequent actions based on environment feedback, have been applied in many domains, including web navigation~\citep{nakano2021webgpt, cheng2024seeclick, he2024webvoyager}, software engineering~\citep{qian2023communicative, hong2023metagpt}, and robotics~\citep{wang2024large, mahadevan2024generative}.
By synergizing the reasoning and action abilities of LLMs, ReAct~\citep{yao2022react} incorporates environment feedback into reasoning traces and determines the next step action dynamically. Subsequent research focuses on integrating code~\citep{wang2023mint, roziere2023code, xu2023lemur}, memory modules~\citep{rana2023sayplan, Park2023GenerativeAgents}, experience reflection~\citep{shinn2023reflexion, zhao2023expel}, and tools into LLM-based agents~\citep{liu2023llm+, patil2023gorilla, qin2023toolllm}, thereby augmenting their abilities in solving complex problems.
However, current LLM-based agents still perform poorly on some complex tasks. This work aims to introduce human interventions and enable humans and agents to collaboratively address complex tasks, thereby achieving improved task performance.


\paragraph{Human-Agent Collaboration}~
In Human-Agent Collaboration (HAC), traditional research has been centered on improving the naturalness and efficiency of human interactions with intelligent agents like robots and AI systems, effectively meeting human needs~\cite{wang2021putting, wu2022survey}. The rise of large-scale language models (LLM-based agents) marks a significant shift in the field, underscoring the role of human feedback and reasoning in enhancing agent capabilities. This approach leverages human insights to refine performance and decision-making processes. Recent studies employ heuristic rules to direct these agents towards seeking human assistance~\cite{cai2023human, wu2022ai, mehta2023improving}. Furthermore, there is an increasing emphasis on developing specialized prompts that motivate LLM-based agents to proactively seek human input, thus nurturing a more interactive and collaborative dynamic in these partnerships~\cite{huang2022inner, wang2023mint}. However, the effectiveness of these methods relies on designing high-quality rules or prompts. This is highly dependent on the designer's domain knowledge. Poor design may result in a system that cannot accurately understand or respond to complex task requirements.  Our research focuses on  designing a generalised and learnable method that coordinates human to effectively work with LLM-based agents in the form of direct planning.

%% file: sections/conclusion.tex
In this paper, we propose the problem of large language model-based human-agent collaboration, delving into the synergy of human intuition and expertise with the computational prowess of LLM-based agents, particularly emphasizing their application in intricate decision-making tasks. We introduce a reinforcement learning-based approach for human-agent collaboration, named \ourmodel. Central to ReHAC is a learnable policy model designed to pinpoint the most critical junctures for human intervention within the task-solving trajectory. Our experimental results show that ReHAC aspects better results and is more generalizable than heuristic rule-based or prompt-based approaches in human-agent collaboration tasks.  We believe that \ourmodel offers a practical pathway for the application of llm-agents in real-world scenarios.


%% file: sections/appendix.tex
\subsection{Experimental Details}
\label{implementation deltails appendix}
\paragraph{Model Implementation}~In our most experiments, we use Llama-2-7b-hf\footnote{\href{https://huggingface.co/meta-llama/Llama-2-7b-hf}{https://huggingface.co/meta-llama/Llama-2-7b-hf}} downloaded from Huggingface as our policy model $\pi_{\theta_1}^{collab}$. We also conduct experiments based on Llama-2-13b-hf\footnote{\href{https://huggingface.co/meta-llama/Llama-2-13b-hf}{https://huggingface.co/meta-llama/Llama-2-13b-hf}} model (see Section \ref{sec:model_scale}). We implement LoRA based on PEFT (\citet{peft}) and set $r_{\text{LoRA}}=16$ and $\alpha_{\text{LoRA}}=16$ for all experiments. Based on \citet{yao2022react} and \citet{yang2023intercode}, we set the step threshold for HotpotQA, StrategyQA, and InterCode to 7, 5, and 8, respectively. All experiments are conducted on NVIDIA A100 GPUs with 40GB memory.

\begin{figure*}[!htbp]
\centering
\includegraphics[width=0.9\textwidth]{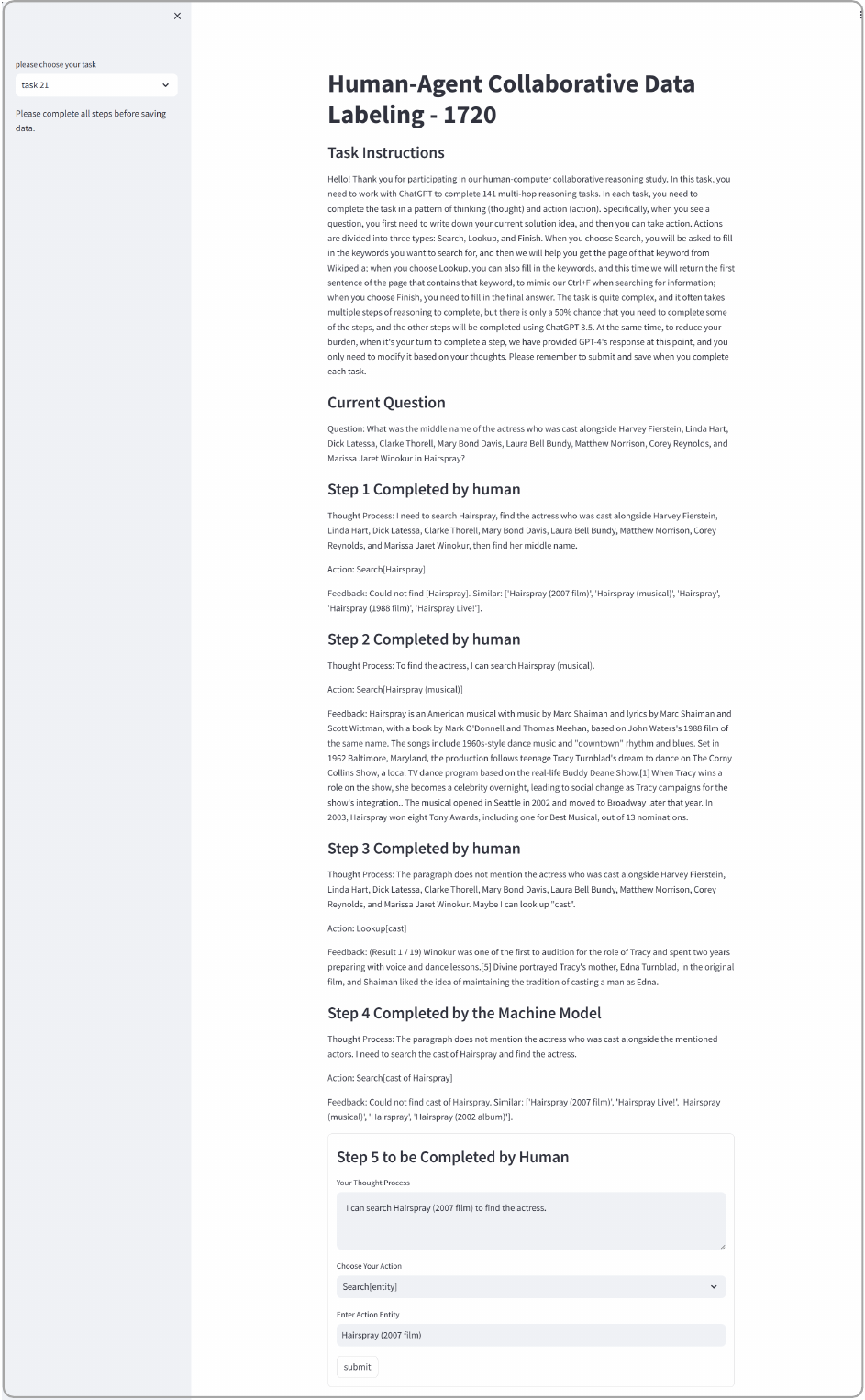} 
\caption{Human-Agent collaborative labelling user interface} 
\label{fig:gui}
\end{figure*}

\paragraph{Human-Agent Dataset}~For a real human-agent collaboration dataset, we employ a uniform sampling method where each action $a_t$ has a 50\% probability of being assigned to either a human annotator or the ChatGPT. 
For each question, we sample as many interaction trajectories as possible. Specifically, for each time $t$, we aim to sample trajectories including $a_t^{collab}=0$ and $a_t^{collab}=1$.
Considering the diversity of responses from different annotators, we permit repeated sampling of the same trajectory during uniform sampling, which means all $a_t^{collab}$ of two trajectories are the same.
To enhance the quality of annotation, annotators are allowed to reference GPT-4's answers.
We recruit 14 annotators through social media, all of whom are graduate students with strong language and reasoning skills. 
They are asked to annotate a total of about 2000 trajectories in four days and they get paid about \$10 an hour. They were explicitly told that the data would be used to train the model and made public and that all the labeled data was unrelated to any individual's privacy.
To facilitate the annotation process, we develop a graphical user interface (GUI)\footnote{The GUI is as shown in Figure \ref{fig:gui}.} and provide one hour of training to annotators. The collected data details are in Table \ref{tab:dataset_detail}. 
\paragraph{GPT-4-Agent Dataset}~For the dataset constructed using GPT-4 to simulate human annotation, we adopt the same sampling method as human-agent dataset collection. However, due to the uniform or near-uniform distribution of GPT-4's responses, we skip duplicate paths during uniform sampling. Collected data details are listed in Table \ref{tab:dataset_detail}.

\subsection{Baselines Details}
\label{baselines deltails appendix}
\paragraph{Random}~We randomly choose a human or an agent to conduct action \(a_t\) at a probability of 50\%.
\paragraph{Prompt}~We prompt an agent to actively decide action \(a_t\) should be finished by itself or a human. The related prompts are shown in Table \ref{tab:prompt method template for QA} and Table \ref{tab:prompt method template for code}.
Experimental results of Random and Prompt are averaged over three repeated experiments.
\paragraph{Imitation Learning}~We select the top 50\% of actions that receive the highest rewards in each state \(s_t\) as expert demonstrations. These expert demonstrations (state-action pairs) are then used to supervise the fine-tuning of the policy model. This approach allows the policy model to learn how to make decisions that get a higher return in a given state.

\begin{table}[!htbp]
  \centering
  \resizebox{\columnwidth}{!}{
  \begin{tabular}{lcccccccccc}
    \toprule
    \multirow{2}{*}{Dataset}&\multicolumn{2}{c}{Trainset}& Testset \\
    \cmidrule(lr){2-3}
    \cmidrule(lr){4-4}
    & Questions & Trajectories & Questions \\
    \midrule
    HotpotQA(real) & 141 & 1937 & 100  \\
    \midrule
    HotpotQA(sim) & 141 & 2135 & 100  \\
    StrategyQA(sim) & 250 & 2420 & 100  \\
    InterCode(sim) & 100 & 2071 & 100 \\
    \bottomrule
  \end{tabular}}
  \caption{Collected dataset details. Questions mean the number of questions we used for human-agent collaboration task. Trajectories mean the overall trajectory number we collected. (real) refers to the real human-agent collaboration dataset, and (sim) refers to the human-agent collaboration dataset collected by using GPT-4 to simulate humans.}
  \label{tab:dataset_detail}
  \vspace{-0.2cm}
\end{table}

\begin{table*}[!htbp]
    \centering
    \resizebox{\columnwidth}{!}{
    \begin{tabular}{lcccccc}
    \toprule
         Experiment & $\alpha$ & $\epsilon$ & Learning Rate & Batch Size \\
         \midrule
         $\text{HotpotQA}_{\lambda=0.06}$(GPT-4-agent, 7b) & 0 & \multirow{10}{*}{0.3} & 3e-5 & \multirow{10}{*}{64} \\
         $\text{HotpotQA}_{\lambda=0.08}$(GPT-4-agent, 7b) & 0 & & 3e-5 \\
         $\text{HotpotQA}_{\lambda=0.10}$(GPT-4-agent, 7b) & 0 & & 5e-5 \\
         $\text{HotpotQA}_{\lambda=0.08}$(GPT-4-agent, 13b) & 0.1 & & 3e-5 \\
         $\text{HotpotQA}_{\lambda=0.06}$(human-agent, 7b) & 0.05 & & 5e-5 \\
         $\text{HotpotQA}_{\lambda=0.08}$(human-agent, 7b) & 0.1 & & 5e-5 \\
         $\text{HotpotQA}_{\lambda=0.1}$(human-agent, 7b) & 0.0 & & 5e-5 \\
         $\text{StrategyQA}_{\lambda=0.08}$(GPT-4-agent, 7b) & 0.1 & & 1e-5 \\
         $\text{InterCode}_{\lambda=0.08}$(GPT-4-agent, 7b) & 0 & & 5e-5 \\
         $\text{InterCode}_{\lambda=0.08}$(GPT-4-agent, 13b) & 0.05 & & 5e-5 \\
         \bottomrule
    \end{tabular}}
    \caption{Hyper-parameter settings for all experiments.}
    \label{tab:hyperparameter}
\end{table*}

\begin{table*}[htbp]
  \centering
  \renewcommand{\arraystretch}{1.1}
  \resizebox{\textwidth}{!}{
  \begin{tabular}{lcccccccccc}
    \toprule
    \multirow{2}{*}{Methods}&\multicolumn{3}{c}{HotpotQA}&\multicolumn{3}{c}{StrategyQA}&\multicolumn{3}{c}{InterCode} \\
    \cmidrule(lr){2-4}
    \cmidrule(lr){5-7}
    \cmidrule(lr){8-10}
    & HIR (\%) & Task Reward & Reward & HIR (\%) & Task Reward & Reward & HIR (\%) & Task Reward & Reward \\
    \midrule
    Agent-only & 0.0 & 22.39 & 22.39 & 0.0 & 60.00 & 60.00 & 0.0 & 53.00 & 53.00 \\
    Human-only & 100.0 & 54.82 & 23.86 & 100.0 & 68.00 & 43.36 & 100.0 & 73.00 & 33.72 \\
    Random & 50.84 & 42.73 & 27.34 & 49.50 & 65.67 & 53.8 & 50.09 & 66.00 & 44.21 \\
    Prompt & 34.06 & 40.46 & 29.26 & 9.14 & 61.33 & 59.12 & 9.94 & 59.33 & 54.69 \\
    IL & 22.08 & 31.50 & 24.70 & 4.76 & 59.00 & 57.88 & 1.01 & 54.00 & 53.52 \\
    Ours & 51.46 & 46.90 & \textbf{31.38} & 20.47 & 66.00 & \textbf{61.12} & 4.15 & 62.00 & \textbf{60.08} \\
    \bottomrule
  \end{tabular}}
  \caption{$\text{\ourmodel}_\text{GPT-4}$ Human intervention rate (HIR), task reward $T$, and reward $R$ of different methods on GPT-4-agent testsets.}
  \label{tab:main_gpt_result}
  \vspace{-0.2cm}
\end{table*}

\begin{table*}[htbp]
    \centering
    \begin{tabular}{p{1.98\columnwidth}}
    \hline
Imagine you are a clever planner. \\
\\
Given an unfinished trajectory with several steps, your task is to decide whether the next step should be carried out by ChatGPT or a human. This decision should be based on a thoughtful evaluation of the difficulty of the next step and the progress made in the current trajectory. Here are two finished trajectory examples. \\
Example 1: \\
\$\{example1\} \\
Example 2: \\
\$\{example2\} \\
Now please decide whether the next step should be carried out by ChatGPT or a human. Please consider the following factors: \\
1. If the next step is relatively straightforward and well within ChatGPT's capabilities, instruct ChatGPT to proceed with the next step. If the task is deemed challenging or requires human judgment, recommend human intervention. \\
2. If the trajectory has been consistently handled by ChatGPT without notable issues, encourage ChatGPT to continue. If there have been challenges or uncertainties in the trajectory, consider suggesting human involvement for the next step. \\
3. Note that human intervention will significantly increase the cost, so try to balance the accuracy and efficiency. \\
If the next step should be carried out by ChatGPT, return [ChatGPT], otherwise, return [Human]. Only return [ChatGPT] or [Human]. \\
\\
\#Your unfinished trajectory\#: \$\{current trajectory\} \\
\#Your return\#: \\
    \hline
    \end{tabular}
    \caption{The prompt template used for the prompt-based method in QA dataset.}
    \label{tab:prompt method template for QA}
\end{table*}

\begin{table*}[htbp]
    \centering
    \begin{tabular}{p{1.98\columnwidth}}
    \hline
Imagine you are a clever planner in SQL. \\
\\
Given an unfinished trajectory with several SQL commands, your task is to decide whether the next command should be carried out by ChatGPT or a human. This decision should be based on a thoughtful evaluation of the difficulty of the next command and the progress made in the current trajectory. Here are two finished trajectory examples. \\
Example 1: \\
\$\{example1\} \\
Example 2: \\
\$\{example2\} \\
Now please decide whether the next command should be carried out by ChatGPT or a human. Please consider the following factors: \\
1. If the next command is relatively straightforward and well within ChatGPT's capabilities, instruct ChatGPT to proceed with the next command. If the task is deemed challenging or requires human judgment, recommend human intervention. \\
2. If the trajectory has been consistently handled by ChatGPT without notable issues, encourage ChatGPT to continue. If there have been challenges or uncertainties in the trajectory, consider suggesting human involvement for the next command. \\
3. Note that human intervention will significantly increase the cost, so try to balance the accuracy and efficiency. \\
If the next command should be carried out by ChatGPT, return [ChatGPT], otherwise, return [Human]. Only return [ChatGPT] or [Human]. \\
\\
\#Your unfinished trajectory\#: \$\{current trajectory\} \\
\#Your return\#: \\
    \hline
    \end{tabular}
    \caption{The prompt template used for the prompt-based method in InterCode dataset.}
    \label{tab:prompt method template for code}
\end{table*}

\subsection{Human Intervention Rate}
\label{human_intervnetion_rate_formula}
We denote the number of steps completed by humans and agents in the dataset by $num_{h}$ and $num_{a}$, respectively. The Human Intervention Ratio (HIR) is calculated as 
\begin{equation*}
    \text{HIR} = \frac{num_{h}}{num_{h}+num_{a}}.
\end{equation*}
HIR measures the rate of human intervention. Generally, a higher HIR indicates better task performance, but it also tends to increase costs.

